\documentclass[conference]{IEEEtran}

\usepackage{cite}
\usepackage{amsmath, amssymb, amsfonts}
\usepackage{textcomp}
\usepackage{booktabs}
\usepackage[nopatch=eqnum]{microtype}
\usepackage[dvipsnames]{xcolor}
\usepackage[colorlinks=true, allcolors=blue]{hyperref}
\usepackage[pass]{geometry}
\usepackage{lipsum}

\usepackage{graphicx}
\usepackage{subfigure}

\usepackage{algorithm}
\usepackage{algpseudocode}
\makeatletter
\renewcommand{\ALG@name}{Prompt}
\makeatother

\usepackage{csvsimple}
\usepackage{listings}
\def\BibTeX{{\rm B\kern-.05em{\sc i\kern-.025em b}\kern-.08em
    T\kern-.1667em\lower.7ex\hbox{E}\kern-.125emX}}

\DeclareMathOperator{\emb}{emb}

\begin{document}
\pagestyle{plain}

\title{Zero-Shot Recommendations with Pre-Trained Large Language Models for Multimodal Nudging}

\author{
    \IEEEauthorblockN{Rachel M. Harrison}
    \IEEEauthorblockA{\textit{Behavioral Design}\\
    \textit{Lirio, LLC}\\
    Knoxville, TN, USA\\
    rharrison@lirio.com}
\and
    \IEEEauthorblockN{Anton Dereventsov}
    \IEEEauthorblockA{\textit{Lirio AI Research}\\
    \textit{Lirio, LLC}\\
    Knoxville, TN, USA\\
    adereventsov@lirio.com}
\and
    \IEEEauthorblockN{Anton Bibin}
    \IEEEauthorblockA{\textit{Skoltech Agro}\\
    \textit{Skoltech}\\
    Moscow, Russia\\
    a.bibin@skoltech.ru}
}

\maketitle

\begin{abstract}
We present a method for zero-shot recommendation of multimodal non-stationary content that leverages recent advancements in the field of generative AI.
We propose rendering inputs of different modalities as textual descriptions and to utilize pre-trained LLMs to obtain their numerical representations by computing semantic embeddings.
Once unified representations of all content items are obtained, the recommendation can be performed by computing an appropriate similarity metric between them without any additional learning.
We demonstrate our approach on a synthetic multimodal nudging environment, where the inputs consist of tabular, textual, and visual data.
\end{abstract}

\begin{IEEEkeywords}
zero-shot learning, large language models, GPT, nudging, personalization, multimodal recommendation
\end{IEEEkeywords}

\section{Introduction}

In this work we propose a method for zero-shot recommendations of multimodal content by utilizing a pre-trained large language model (LLM).
We demonstrate our approach on a simulated nudging task in the form of a screen time management application that sends tailored notifications consisting of a message and an image when the user reaches their prescribed screen time limit.
The particular difficulty of such a setting lies in the \textit{cross-modality} (any message can be combined with any image) and \textit{non-stationarity} (the set of available messages and images can be changed at will) of the content.
We further explore the intricacies of zero-shot learning within the framework of LLMs, shedding light on the innovative capabilities it brings to the analysis of textual descriptions for formerly non-textual content, and ultimately contributing to the advancement of multimodal recommendation and nudging.

\begin{figure*}[t]
    \centering
    \includegraphics[width=.95\linewidth]{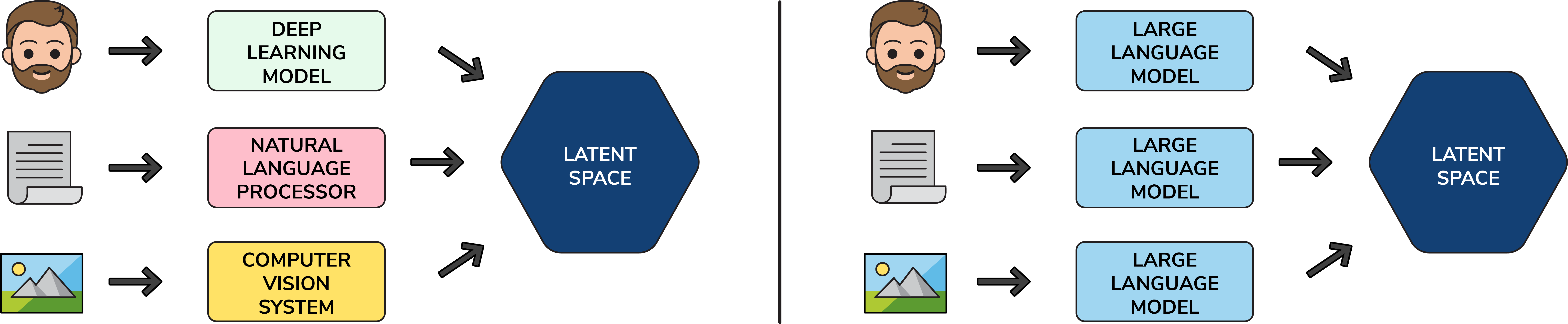}
    \caption{A conventional approach to multimodal recommendation (left) and the proposed zero-shot approach (right).}
    \label{fig:ml_vs_zs}
\end{figure*}

Given the recent developments in the fields of natural language processing and generative AI, we believe that the use of pre-trained LLMs can help practitioners obtain appropriate input representations instead of learning an encoder for each modality from scratch (see Figure~\ref{fig:ml_vs_zs}), thus accelerating the development of real-world nudging and hyperpersonalization applications.
Our main contributions are as follows:
\begin{itemize}
    \item Design a highly configurable synthetic nudging environment with non-stationary multimodal content items;
    \item Propose a method for obtaining a unified numerical representation for inputs of different modalities;
    \item Describe an approach for zero-shot recommendation in a multimodal setting using a pre-trained LLM.
\end{itemize}

\begin{figure}[t]
    \centering
    \includegraphics[width=.93\linewidth]{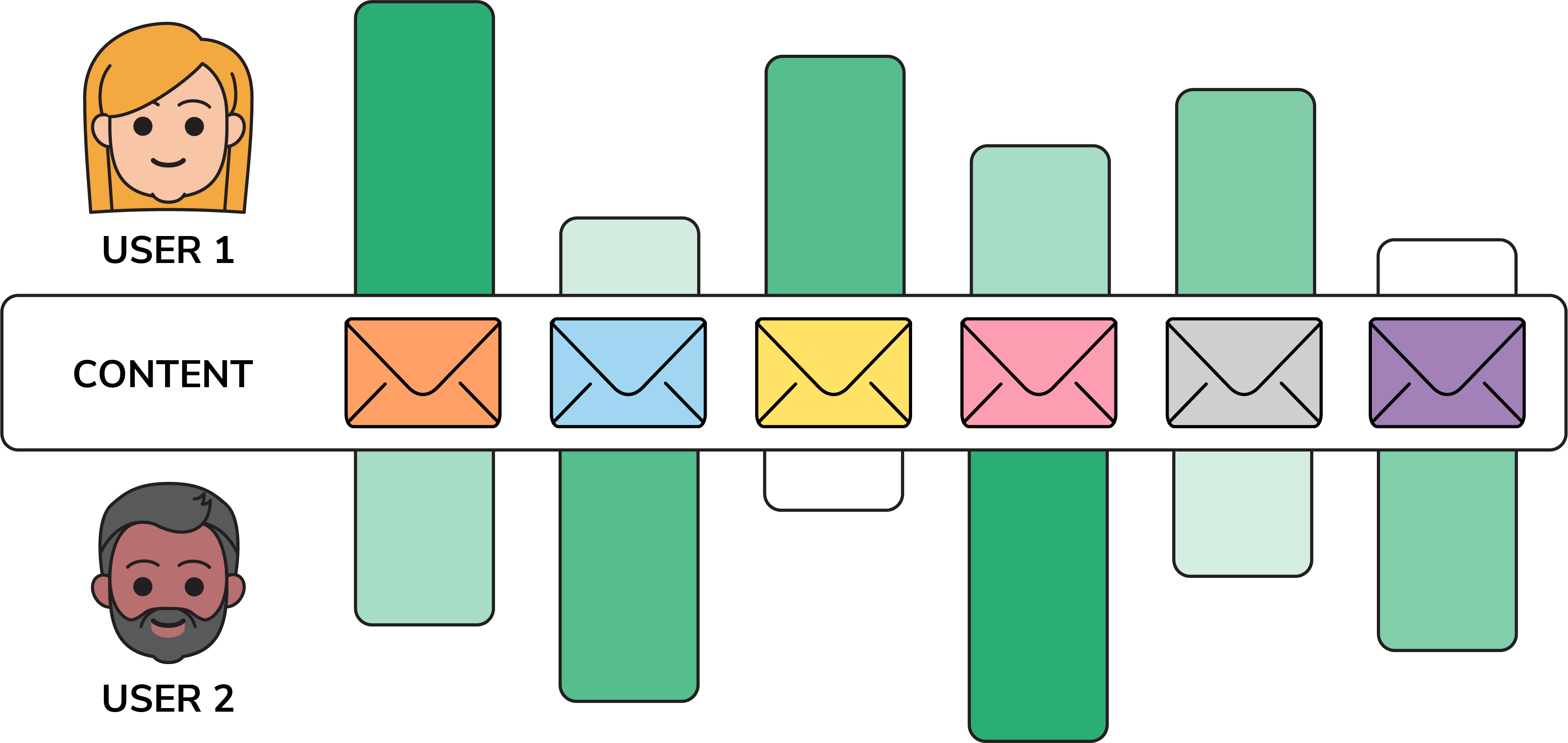}
    \caption{A general content recommendation setting.}
    \label{fig:content_recommendation}
\end{figure}

\subsection{Machine Learning for Recommendation and Nudging}

In recent years, the rapid growth of digital platforms and the abundance of online content have underscored the importance of efficient and personalized recommendations~\cite{nabizadeh2020learning}.
Recommender systems, often powered by machine learning techniques, play a pivotal role in guiding users through an overwhelming sea of choices by suggesting items, services, or content tailored to their preferences, as illustrated in Figure~\ref{fig:content_recommendation}.
Leveraging vast amounts of user and content data via intricate algorithms, machine learning enables these systems to discern intricate patterns and preferences that traditional rule-based approaches would struggle to capture, see~\cite{portugal2018use, jesse2021digital, lex2021psychology}.

As recommender systems have matured, their objectives have expanded beyond mere utility; they now aim to deeply understand user behaviors and preferences to enhance user engagement and satisfaction.
Furthermore, this evolution has given rise to the integration of behavioral economics principles like ``nudging'' into recommendation engines.
Nudging involves strategically influencing user decisions without restricting their autonomy~\cite{thaler2009nudge}.
Consider the example of video streaming platforms: machine learning algorithms analyze a user's historical viewing habits and preferences to suggest relevant movies or shows.
By incorporating nudging techniques, such as placing highly-rated content prominently on the homepage, users are more likely to be presented with and guided towards new choices that align with their existing preferences.
As a result, nudging via recommender systems facilitate more personalized content for individual users, thereby enhancing the probability of user engagement and satisfaction with the suggested content.


\subsection{Multimodal Recommender Systems}

While traditional approaches to recommender systems often focus on delivering a singular type of content to the user, multimodal systems combine various data sources like text, images, user interactions, and contextual data, to construct more intricate and tailored recommendations.
In the case of a video streaming platform, a multimodal system could create a more personalized experienced by not only suggesting movies but also by designing unique title cards and summaries that align with the user's preferences.
This level of customization enhances user engagement and satisfaction by resonating with their individual interests on a granular level.
Such meticulous tailoring gains substantial importance in fields like healthcare personalization, where decisions are intricately connected to behaviors that are notoriously difficult to change~\cite{vlaev2016theory, middleton2013long}.


Despite the many potential benefits of multimodal recommender systems, a significant challenge arises from the inherent difficulty in effectively matching disparate content.
Unlike homogeneous content, multimodal content encompasses various forms of data that are fundamentally distinct in nature.
The task of linking an image to a relevant piece of text or associating a video with its corresponding title becomes challenging due to the disparities between these modalities.
This challenge has posed a historical hindrance to the seamless integration of diverse forms of data into recommendation systems, as, while the possible advantages are substantial, the dissimilarity between modalities necessitates sophisticated solutions to enable effective cross-modal comparisons, see e.g.~\cite{zhou2023comprehensive, liu2023multimodal}.
Addressing this challenge is crucial for unlocking the full potential of multimodal recommender systems and providing users with recommendations that cater to their preferences across various dimensions of content.

\subsection{Zero-Shot Learning with LLMs}

Zero-shot learning offers a compelling strategy to enhance the capabilities of recommendation systems in unprecedented ways.
Unlike conventional approaches that require exhaustive training on each modality, zero-shot learning equips models with the ability to generalize from existing knowledge to new and unknown data~\cite{wang2019survey, pourpanah2022review}.
By integrating zero-shot learning with large language models within the realm of multimodal recommendation, these systems can leverage the inherent language understanding prowess of LLMs to decipher nuanced user preferences and content characteristics across various modalities.
Crucially, this approach circumvents the need for exhaustive training on each modality, which can be both time-consuming and resource-intensive.
Instead, by capitalizing on the wealth of pre-trained knowledge within the LLM, the recommendation system can rapidly adapt to new modalities and user preferences.
The integration of zero-shot learning with LLMs thus has the potential to provide users with more tailored recommendations, incorporating diverse modalities without compromising on accuracy or coverage~\cite{lin2023can, fan2023recommender}.

\subsection{Related Work}

In our setting, the inputs (users, messages, and images) are given in different modalities.
As such, our task is related to other multimodal recommender systems, e.g.~\cite{truong2019multimodal, salah2020cornac, wu2021mm}.

We design a synthetic screen time app to illustrate how different components can be used to motivate behavior change in users, which is in the area of digital nudging~\cite{karlsen2019recommendations, shmakov2021nudge, dalecke2020designing}.

Content-matching machine learning techniques enable a personalized digital environment for a given user based on their demographic information and preferences.
Such an approach is becoming increasingly popular, especially in the areas of healthcare interventions and personalization~\cite{zhu2018robust, hassouni2018personalization, gomez2022whom}.

We propose a zero-shot approach to compute user--content preferences, which is similar to~\cite{li2019zero, ding2021zero}.
Moreover, we are employing a pre-trained LLM to extract semantic representations for different inputs, which is similar to zero-shot recommendation approaches proposed in~\cite{wang2023zero, hou2023large}.

Our simulated environment can be utilized as a benchmark for personalization and nudging methods, thereby addressing an existing gap in the community, see e.g.~\cite{rohde2018recogym, ie2019recsim, dereventsov2021unreasonable, dereventsov2022simulated}.

\begin{table*}[t]
    \centering\footnotesize
    \caption{Example of users generated via GPT-4.}
    \label{tab:generated_users}
    \begin{filecontents*}{./data/users_sample.csv}
,Gender,Age,Race,Likes,Dislikes
0,Female,23,White,{active, outdoors, learning},{mental, crafts, homemaking}
1,Male,42,Black,{outdoors, mental, active},{learning, exploration, arts}
2,Female,25,Asian,{crafts, indoors, arts},{passive, mental, outdoors}
3,Male,68,Hispanic,{passive, indoors, homemaking},{relaxation, learning, physical}
4,Female,39,White,{mental, homemaking, crafts},{outdoors, active, relaxation}
5,Male,43,Black,{outdoors, exploration, relaxation},{indoors, passive, homemaking}
6,Female,22,Hispanic,{active, mental, homenaking},{learning, exploration, relaxation}
7,Male,45,Asian,{indoors, passive, exploration},{outdoors, active, mental}
8,Female,35,White,{active, outdoors, exploration},{passive, indoors, physical}
9,Male,31,Hispanic,{mental, learning, relaxation},{outdoors, active, arts}
\end{filecontents*}
\csvautobooktabular{./data/users_sample.csv}
\end{table*}

\section{Problem Setting}\label{sec:problem_setting}

In order to illustrate the potential of our multimodal nudging approach for matching diverse types of content across a wide range of user demographics, we have chosen the setting of screen time monitoring. 
According to recent data, the average person spends 7 hours per day on smartphones, tablets, computers, and other electronic devices connected to the internet~\cite{datareportal2022}.
Increased sedentary screen time has been associated with lowered physical and psychological well-being~\cite{davies2012associations, twenge2019more} with links to a number of health issues like type 2 diabetes~\cite{hu2003television}, obesity~\cite{hu2003television}, and depression~\cite{madhav2017association, hamer2013television, de2011sedentary}.
This widespread contemporary issue impacting individuals across all demographics offers an ideal backdrop for showcasing the potential of our strategy for customized nudging.
By using screen time notifications encouraging participation in offline activities as an example use case, we are able to illustrate how our approach caters to a diverse population with varying types of multimodal inputs, thus reflecting the breadth of its applicability in numerous real-world applications.


To replicate the types of user and content data that might be used for crafting customized screen time notifications, 
our inputs are given in three different modalities: tabular (users), text (messages), and visual (images).
We construct a dataset consisting of 20 users, 40 messages, and 50 images (for examples see Tables~\ref{tab:generated_users} and~\ref{tab:generated_messages} and Figure~\ref{fig:generated_images} respectively), which allows for a total of 2000 unique image--message combinations per user. 
The full dataset is included in the Appendix.

\subsection{Data Design and Prompt Engineering}

To reduce bias and ensure both diversity of inputs and public accessibility to our data design process, all data is either generated via publicly available LLMs or obtained from otherwise free-to-use sources.
Generative AI was chosen for the task of content creation due to its ability to simulate the diverse and potentially uncontrollable variety of content one might have available when designing nudging applications and its allowance for transparency into the design process.
Additionally, each input type in our simulation is intentionally designed to vary in structure between the three modalities, once again exemplifying the heterogeneity of data that one would be attempting to match in real-world multimodal applications.

All generations are performed in Jupyter notebooks running Python 3.8 on a consumer-grade laptop.
The process of content generation/selection for each of the three input types is detailed below.
We note that since LLMs have been used with their default parameters, the data generation process is transparent, but potentially not reproducible.
The source code for data the generation and subsequent matching is available online at~\url{https://github.com/paxnea/LLM-multimodal-nudging}.






\subsection{User Generation}

Users are generated using the \texttt{gpt-4-0613} model\footnote{\url{https://platform.openai.com/docs/models/gpt-4}} with default parameters via OpenAI API calls.
Using a single zero-shot user prompt with the default system message, the model is prompted with the request given in Prompt~\ref{prompt:user_generation} to output users as a database similar to what might be used in practical applications.
Specifically, each user is given as a vector consisting of the following attributes: \texttt{Gender}, \texttt{Age}, \texttt{Race}, \texttt{Likes}, and \texttt{Dislikes}.
Likes and dislikes for each user are sampled from 12 broad activity types~--- active, passive, indoors, outdoors, mental, physical, arts, crafts, exploration, relaxation, learning, homemaking~--- which are chosen based on their generality in reference to the more specific activities proposed in the generated messages.

In total, 20 unique users are generated with all required attributes. Examples of generated users are presented in Table~\ref{tab:generated_users}.





\begin{algorithm}[h]
\begin{lstlisting}
You are tasked with generating a CSV-formatted dataset of 20 people. Each person must have a Gender, Age, Race, Likes, and Dislikes. Race must be either White, Black, Asian, or Hispanic. Likes and dislikes should be separated by semicolons.

There are twelve categories of activities that can be either Likes or Dislikes: active, passive, indoors, outdoors, mental, physical, arts, crafts, exploration, relaxation, learning, homemaking.
Each person should have three Likes and three Dislikes, sampled at random.

Generate a CSV-formatted dataset of 20 diverse people with different Gender, Age, Race, Likes, and Dislikes.
\end{lstlisting}
    \caption{Prompt for user generation with GPT-4}
    \label{prompt:user_generation}
\end{algorithm}

\begin{table*}[t]
    \centering\footnotesize
    \caption{Example of messages generated via GPT-3.5.}
    \label{tab:generated_messages}
    \begin{filecontents*}{./data/messages_sample.csv}
,Message
0,{Get up and groove! Dancing is your ticket to fun, fitness, and a break from the screen. Let the music move you!}
1,{Ditch the screen, hit the trail! Discover nature's wonders, boost your mood, and energize your body with an invigorating hike.}
2,{Give your eyes a break and let your body flow with the ancient practice of yoga. Feel the tension melt away and experience the incredible benefits of increased flexibility, strength, and inner peace.}
3,{Hey! Why not leave the screen behind and hop on your bike? Feel the wind in your hair, soak up the sunshine, and boost your mood while getting some exercise. Let's go biking and enjoy the great outdoors!}
4,{Step away from the screen and let's stroll! Walking boosts energy, clears the mind, and fills your day with fresh inspiration. Let's get moving!}
5,{Dive into the pool and feel the freedom! Swap screens for a refreshing swim, unleash your energy, and let the water ignite your senses.}
6,{Why not give your thumbs a break and start pumping some iron? Weightlifting will boost your strength, tone your muscles, and leave you feeling unstoppable!}
7,{Hey! Ready to get moving? Put down the screen and let's hit the pavement - jogging boosts energy, reduces stress, and gives you a natural high! Let's go!}
8,{Step away from the screen and let's roll! Experience the thrill of roller skating - it's exhilarating, boosts your mood, and keeps you active. Lace up and let the fun begin!}
9,{Step away from the screen and tee off into a world of fun and relaxation! Golfing boosts your mood, improves focus, and gets you outdoors enjoying nature. Let's swing into action!}
\end{filecontents*}
\csvreader[tabular=lp{.9\linewidth},
               table head=\toprule&Message\\\midrule,
               late after last line=\\\bottomrule]
        {./data/messages_sample.csv}{}{\csvcoli&\csvcolii}
\end{table*}

\subsection{Message Generation}

Messages are generated using the \texttt{gpt-3.5-turbo-0613} model\footnote{\url{https://platform.openai.com/docs/models/gpt-3-5}} with the default parameters via OpenAI API calls.
Messages are designed to be concise but specific, with each message focusing on encouraging the user to reduce their screen time by engaging in a particular offline activity instead.
Due to the complexity of linguistic and structural requirements for messages, message generation consists of a system prompt and a user prompt, given in Prompt~\ref{prompt:message_generation}.

In total, 40 unique messages are generated, each representing a specific offline activity.
Each activity corresponds to one or more general activity types available as Likes or Dislikes in the user dataset. 
Examples of the generated messages are presented in Table~\ref{tab:generated_messages}.



\begin{algorithm}[h!]
\begin{lstlisting}[morekeywords={SYSTEM, USER}]
SYSTEM: You are writing a message encouraging users to reduce their screen time.
The message will be delivered when the user has exceeded the amount of recommended screen time, and should encourage the recipient to step away from their screen and engage in {activity} instead.
The message should: be concise -- no longer than two short sentences, use simple language so everyone can understand, be friendly, fun, positive, enthusiastic, and motivating, directly reference {activity} and make it sound enticing, mention the benefits of {activity}.
The message should NOT: be overly wordy, contain greetings like "hey there!" or "hey!", use the words "refreshing," "time," "escape," "swap," take a break," or "unleash your creativity", use emojis or be encased in quotation marks.
The goal is to have the recipient put down their phone and be inspired to start
{activity} immediately.

USER: Generate a motivating message that proposes the user engage in {activity}.
\end{lstlisting}
    \caption{Prompts for message generation with GPT-3.5}
    \label{prompt:message_generation}
\end{algorithm}

\subsection{Image Selection}

Images are selected from the royalty-free stock photo gallery Pexels\footnote{\url{https://www.pexels.com/}} based on their relevance to the setting.
Our selection consists of images that either portray obvious depictions of activities referenced in the generated messages (e.g. biking, reading, cleaning), or are generic yet versatile displays of ambiguous activities (e.g. someone outdoors looking up at the sky, someone indoors sitting at a desk).
Each photo contains a person engaging in some sort of activity, and the face of the person is visible in 80\% of the photos to allow for better identification of the subject's age, race, and gender.
For standardization of sizes and resolutions, all images are resized to 512 px $\times$ 384 px at 72 ppi.

\begin{figure*}
    \centering
    \subfigure[senior, white, male, holding a book, walking]{\includegraphics[width=.19\linewidth]{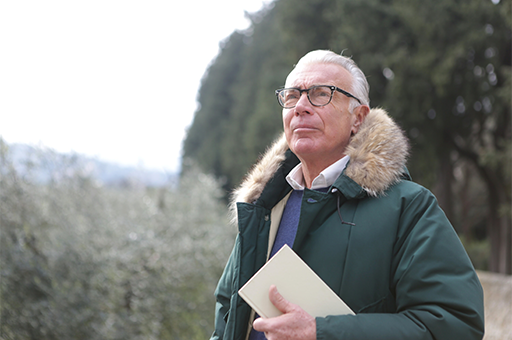}}
    \subfigure[male, young, adult, and he is tying his shoe]{\includegraphics[width=.19\linewidth]{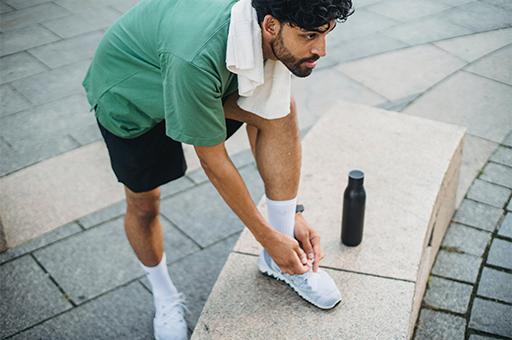}}
    \subfigure[woman, asian, adult, watering plants, watering plants]{\includegraphics[width=.19\linewidth]{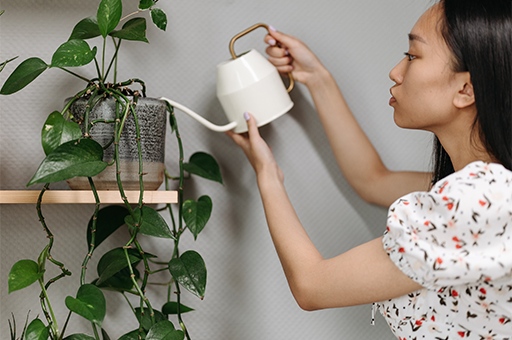}}
    \subfigure[woman, adult, white, sitting at a desk, writing]{\includegraphics[width=.19\linewidth]{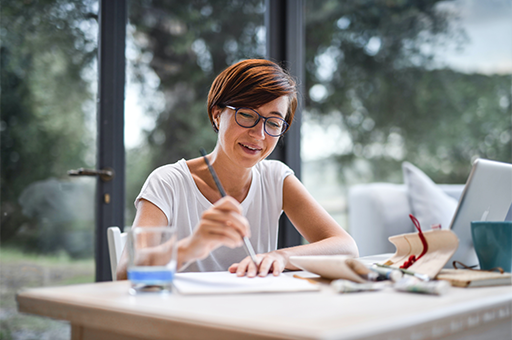}}
    \subfigure[woman, adult, afro-american, meditating, yoga]{\includegraphics[width=.19\linewidth]{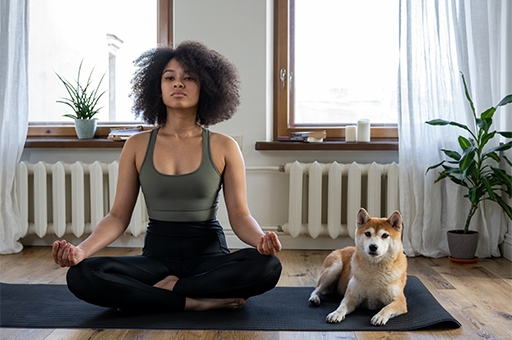}}
    \caption{Examples of images selected from Pexels and the corresponding captions generated via the BLIP-2 framework.}
    \label{fig:generated_images}
\end{figure*}

In total, 50 images were chosen to be paired with the generated messages and provide reasonable demographic diversity in their depictions of subjects.
Examples of the selected images are presented in Figure~\ref{fig:generated_images}.


\section{Methods}\label{sec:methods}

Essentially, the main idea of our approach is to describe each input (regardless of modality) by a string of text and leverage a pre-trained LLM to obtain a relevant numerical representation, which is then used to perform content matching and recommendation.
Below we describe in detail how these steps are performed in our setting; however, in practice, the exact realization of these steps could be tailored to suit each specific application.

\subsection{Inputs Representation}

Before matching can take place, a unified numerical representation for each input must be obtained by embedding all content items in the same latent space.
In essence, our content representation consists of describing each input by a string(s) of text, using a pre-trained LLM to compute embeddings for each text description, and adjusting these embeddings to obtain a relevant representation.
While assigning text descriptions to inputs depends heavily on the specific application, we present an example of a possible approach.

Let $\emb : \mathcal{T} \to \mathbb{R}^d$ be the embedding function of a given LLM with $\mathcal{T}$ denoting the set of all text strings and $d$ denoting the dimensionality of the embedding space.
Below we describe how, with the use of an embedding function $\emb$, each type of content is represented in our setting.

\subsubsection{Message Embeddings}
Let $\mathcal{M}$ denote the set of available messages.
Since each message $m \in \mathcal{M}$ is given as a string of text, we simply compute its representation $r(m)$ as
\[
    r(m) = \emb(m) \in \mathbb{R}^d.
\]

\subsubsection{Image Embeddings}
Let $\mathcal{I}$ denote the set of available images.
In order to represent an image $i \in \mathcal{I}$ as a string of text, we generate captions using the \texttt{BLIP-2} framework~\cite{li2023blip} with the pre-trained \texttt{flan-t5-xxl} LLM\footnote{\url{https://huggingface.co/Salesforce/blip2-flan-t5-xxl}}, used via Hugging Face Inference API\footnote{\url{https://huggingface.co/docs/transformers/main/model_doc/blip-2}}.
The captioning is performed via prompted visual question answering (Prompt~\ref{prompt:image_captioning}).
The model is specifically prompted to give information on the age, race, and gender of the subject in the photos, as well as a description of the activity being performed.
See Figure~\ref{fig:generated_images} for examples of the generated captions.
After each caption is obtained for a given image $i \in \mathcal{I}$, an embedding is computed, which serves as a representation $r(i)$ of the image $i$, i.e.
\[
    r(i) = \emb(\textrm{caption}(i)) \in \mathbb{R}^d.
\]

\begin{algorithm}[h]
\begin{lstlisting}
Question: What is the gender, age (young, adult, senior), and race of the person in this photo, and what are they doing? Answer:
\end{lstlisting}
    \caption{Prompt for image captioning with BLIP-2}
    \label{prompt:image_captioning}
\end{algorithm}

\subsubsection{User Embeddings}
Let $\mathcal{U}$ denote the set of available users, which are defined by their demographic features and their liked/disliked activities (Table~\ref{tab:generated_users}).
For a user $u \in \mathcal{U}$ we first convert the demographic information into a sentence and compute its embedding.
Then for each like/dislike we compute a corresponding embedding and add/subtract it from the demographic embedding with a prescribed weight.
The process of computing a representation $r(u)$ is the following:
\begin{align*}
    r(u)
    &= \emb(\textrm{demographics}(u))
    \\
    &+ w_l \sum_{a \in L} \emb(a)
    - w_d \sum_{a \in D} \emb(a)
    \in \mathbb{R}^d,
\end{align*}
where $L$ and $D$ denote the liked/disliked activities for the current user.
Parameters $w_l$ and $w_d$ control the effect the user's preferences will have on the provided recommendations.
For instance, setting larger values for $w_l$ and $w_d$ is suitable for more risk-averse programs where precise alignment with the user preferences is crucial.
In our experiments, we use the values $w_l = w_d = 0.2$.
For example, we compute the representation $r(u)$ for the first user $u$ from Table~\ref{tab:generated_users} as follows:
\begin{align*}
    r(u) &= \emb(\textrm{``23 year old white female''})
    \\
    &+ 0.2 \emb(\textrm{``active''})
    + 0.2 \emb(\textrm{``outdoors''})
    \\
    &+ 0.2 \emb(\textrm{``learning''})
    - 0.2 \emb(\textrm{``mental''})
    \\
    &- 0.2 \emb(\textrm{``crafts''})
    - 0.2 \emb(\textrm{``homemaking''})
\end{align*}

\begin{figure*}[t]
    \centering
    \hfill
    \includegraphics[width=.45\linewidth]{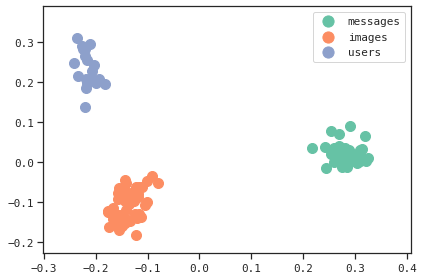}
    \hfill
    \includegraphics[width=.45\linewidth]{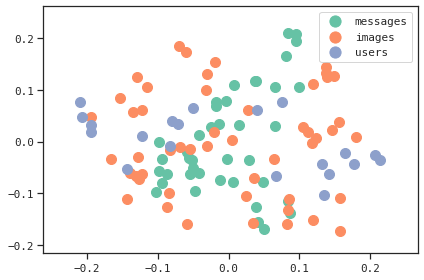}
    \hfill
    \caption{Distribution of vector representations: PCA-projection of un-centered (left) and centered (right) input representations.}
    \label{fig:embeddings_centring}
\end{figure*}

\subsection{Preference Matching}

In this work we use the \texttt{text-embedding-ada-002} model\footnote{\url{https://platform.openai.com/docs/models/embeddings}} to compute the embeddings.
Due to the specifics of this particular model, for any string of text $t \in \mathcal{T}$ the embedding operator $\emb$ satisfies
\[
    \emb(t) \in \mathbb{S}^{1535} \subset \mathbb{R}^{1536},
\]
where $\mathbb{S}^{1535}$ denotes the unit sphere in $\mathbb{R}^{1536}$.
As a result, the computed input representations are clustered by input type, see Figure~\ref{fig:embeddings_centring} (left).
Evidently, such a skewed distribution complicates the process of content matching.
To address this issue, we ``normalize'' the input representations by offsetting each representation vector by the center of its cluster, see Figure~\ref{fig:embeddings_centring} (right).
Thus, for an input $x \in \mathcal{X}$ the ``normalized'' representation $\bar{r}(x)$ is computed as
\[
    \bar{r}(x) = r(x) - \frac{1}{|\mathcal{X}|} \sum_{x^\prime \in \mathcal{X}} r(x^\prime),
\]
where $\mathcal{X} \in \{\mathcal{M}, \mathcal{I}, \mathcal{U}\}$ represents the set of all inputs of a given type (i.e. messages, images, or users).
Note that such an offsetting allows each input to affect the representation of every other item of the same type, which can be viewed as a (rather trivial) realization of collaborative filtering~\cite{su2009survey}.

For a given user $u \in \mathcal{U}$, a message $m \in \mathcal{M}$, and an image $i \in \mathcal{I}$ the preference of a tuple $(u, m, i)$ is computed as
\begin{equation}\label{eq:preference}
    \begin{aligned}
        p(u, m, i)
        &= w_{m,i} \big\langle \bar{r}(m), \bar{r}(i) \big\rangle,
        \\
        &+ w_{u,m} \big\langle \bar{r}(u), \bar{r}(m) \big\rangle
        + w_{u,i} \big\langle \bar{r}(u), \bar{r}(i) \big\rangle
    \end{aligned}
\end{equation}
where $w_{m,i}, w_{u,m}, w_{u,i}$ are configurable parameters controlling the importance of the alignment of each pair of content items, and $\langle \cdot, \cdot \rangle : \mathbb{R}^{1536} \times \mathbb{R}^{1536} \to \mathbb{R}$ denotes the inner product.
The values of $w_{m,i}, w_{u,m}, w_{u,i}$ can be tailored for the practical needs of each particular application.
For simplicity, in our case we set $w_{m,i} = w_{u,m} = w_{u,i} = 1$.

The rationale of computing the preference $p(u,m,i)$ via~\eqref{eq:preference} is validated in that the compatibility of each pair of inputs $(m,i), (u,m), (u,i)$ contributes to the overall preference score for the tuple $(u,m,i)$.
Moreover, controlling the weights $w_{m,i}, w_{u,m}, w_{u,i}$ allows one to accentuate the matching between different input types in accordance with the given application.

To provide $k$ recommendations for a given user $u \in \mathcal{U}$, compute the preferences over all possible pairs $(m, i)$, i.e.
\[
    \mathcal{P}(u) = \big\{ p(u, m, i) \,\big|\, m \in \mathcal{M}, i \in \mathcal{I} \big\}
\]
and select $k$ pairs $(m, i)$ corresponding to the largest values of $p(u,m,i)$.
Alternatively, one can obtain a probability distribution by normalizing the set $\mathcal{P}(u)$ (via e.g. softmax) and sampling content recommendations $(m,i)$ from the resulting distribution, which naturally allows for the deployment of reinforcement learning agents, see e.g.~\cite{sutton2018reinforcement}.

\section{Results}

We deploy our approach as explained in Section~\ref{sec:methods} on the simulated task defined in Section~\ref{sec:problem_setting} and construct tailored multimodal content recommendations for users.
Specifically, for each user $u \in \mathcal{U}$ we rank all possible recommendations $(m,i)$, consisting of two items: a message $m \in \mathcal{M}$ and an image $i \in \mathcal{I}$.
To assess the quality of a recommendation $(m,i)$ for a user $u$, we first check the validity of the following conditions:
\begin{enumerate}
    \item the message $m$ is aligned with the image $i$~--- the activity referenced in the message is either shown explicitly or can be assumed from the image--message pair;
    \item the message $m$ agrees with the user's $u$ likes~--- the activity referenced in the message aligns with at least one of the user's likes;
    \item the image $i$ matches the user's $u$ demographics~--- the person in the image matches at least two of the user's following characteristics: gender, age, race.
\end{enumerate}
We say that a recommendation $(m,i)$ is appropriate for a user $u$ if condition 1 and at least one of conditions 2 and 3 hold.
A recommendation is considered inappropriate if either the message $m$ does not align with the image $i$ or the message $m$ suggests an activity that matches one of the user's $u$ dislikes.

\begin{figure*}
    \centering
    \includegraphics[width=.9\linewidth]{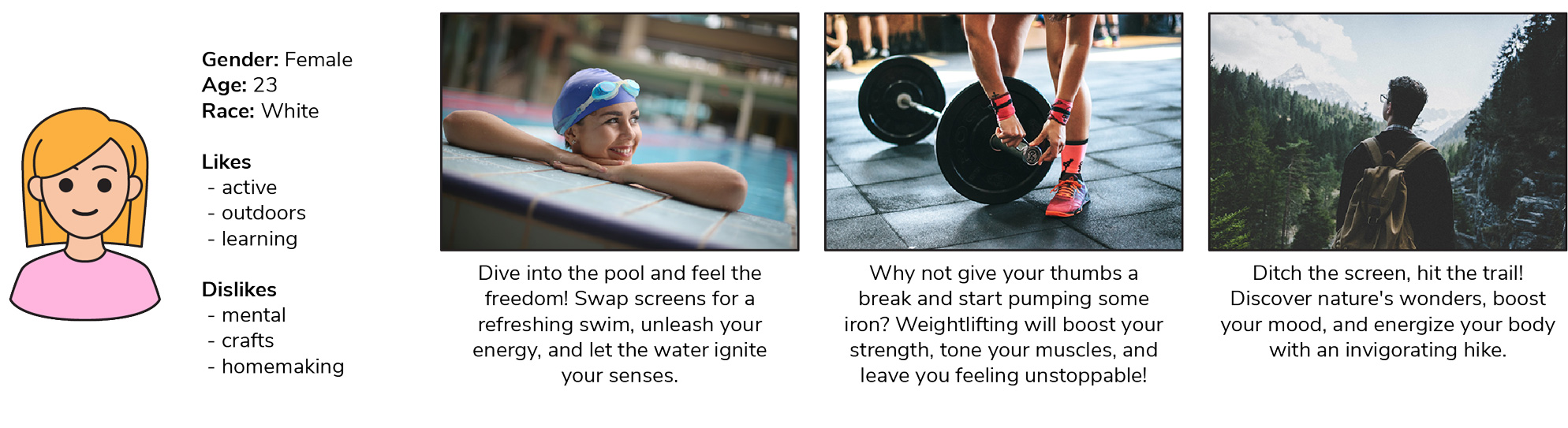}
    \\
    \includegraphics[width=.9\linewidth]{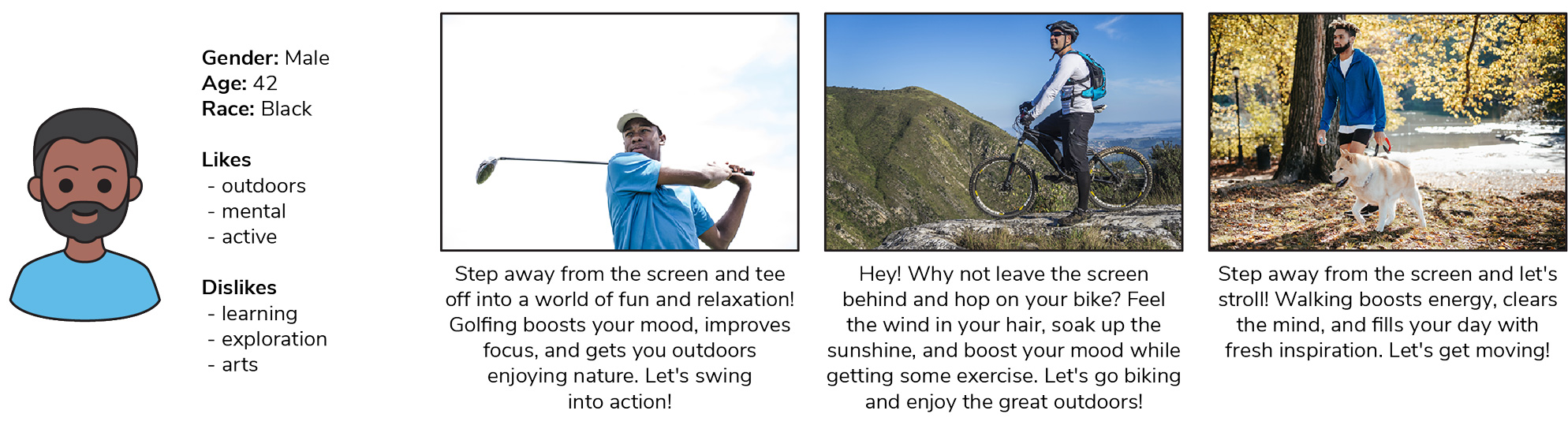}
    \\
    \includegraphics[width=.9\linewidth]{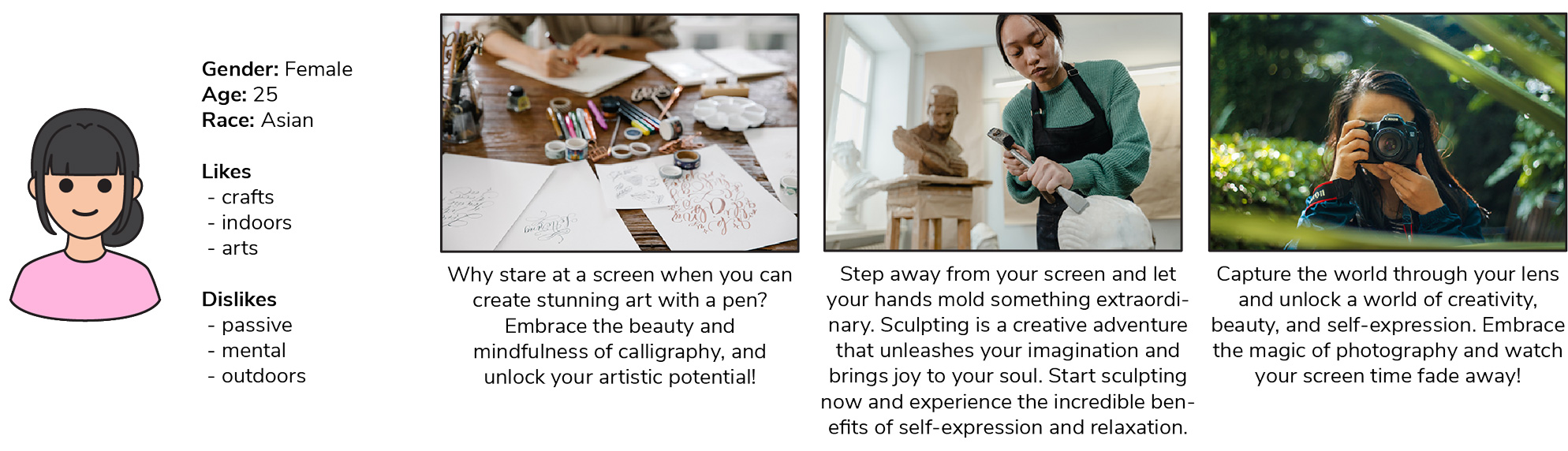}
    \\
    \includegraphics[width=.9\linewidth]{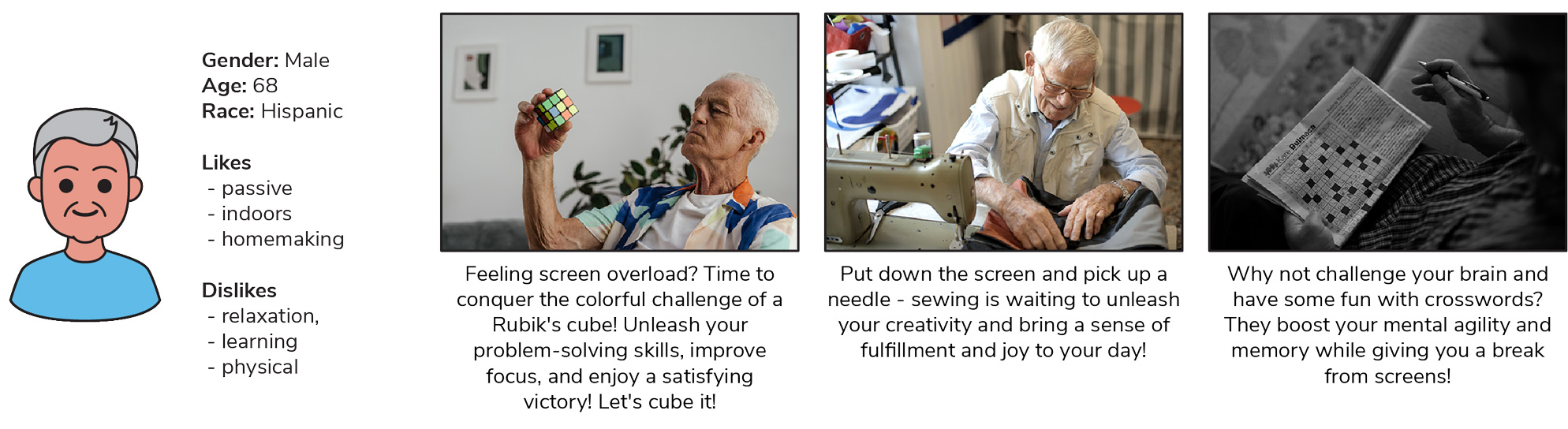}
    \\
    \includegraphics[width=.9\linewidth]{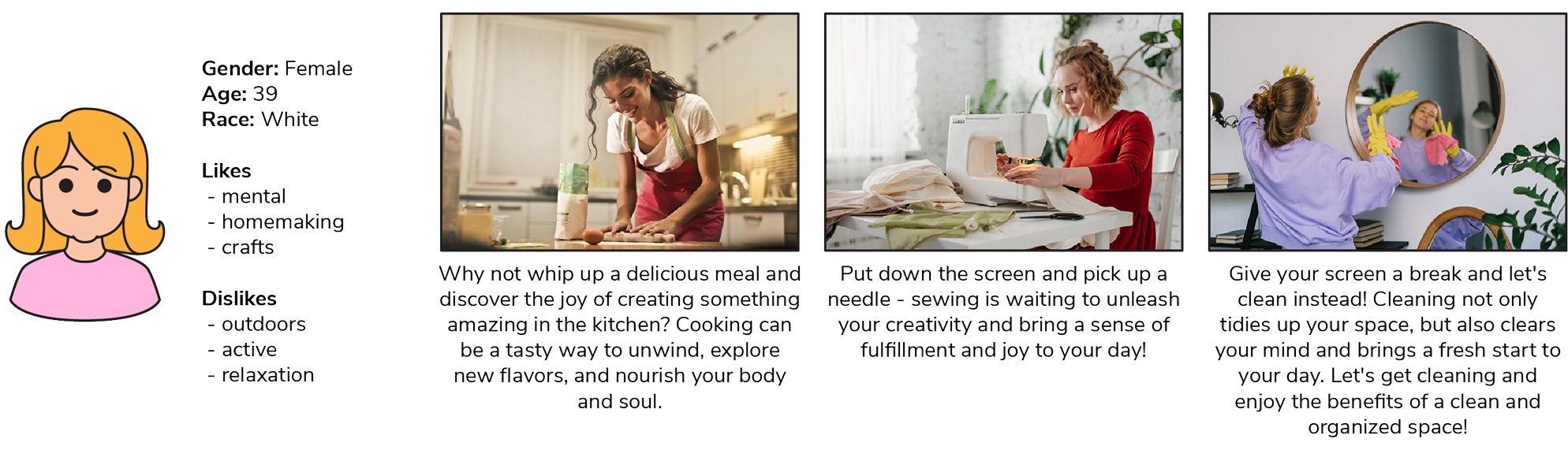}
    \caption{Content recommendations provided by a zero-shot approach: top 3 nudges for each of the first 5 users.}
    \label{fig:results}
\end{figure*}

For each of the 20 users in our dataset we compute the top 5 recommendations (out of the potential 2000 combinations) and evaluate their appropriateness according to the criteria above.
We determine that 83\% of the recommendations fully satisfy the criteria and only 8\% do not.
In Figure~\ref{fig:results} we display the top 3 recommendations for the first 5 users from Table~\ref{tab:generated_users}.

\section{Discussion}

In this work we construct a synthetic multimodal nudging task that necessitates different types of inputs be matched to construct personalized image--message combinations for a given user.
We propose a zero-shot recommendation approach that leverages pre-trained LLMs to convert all inputs to textual descriptions and use their semantic embeddings to obtain unified numerical representations for each content item.
We admit that the presented approach is rather heuristic and is not intended to be treated as a finished product, but rather as an example of the possibilities resulting from recent advancements in generative AI.
Nevertheless, we believe that our results are of strong interest to the community in nudging and hyper-personalization.

As all of our inputs are generated and/or processed with pre-trained LLMs, there exists the potential for unintentional bias in our results if the same LLM is used for both creating and matching the data.
Thus, in order to avoid generation from the same distribution, we employ different LLMs to perform different tasks.
Specifically, we employ \texttt{gpt-4-0613} for user generation, \texttt{gpt-3.5-turbo-0613} for message generation, \texttt{flan-t5-xxl} for image caption generation, and \texttt{text-embedding-ada-002} for embedding calculations.
Using different LLMs ensures that the data matching is highly non-trivial and our results are not biased to a particular LLM.

In our environment, the users' interests are presented in coarse categories, which might not be suitable for some applications; this can be addressed by replacing the interests with more fine-grained activities.
Each interest is represented by its embedding via a given LLM, and is used to shift the user's representation vector.
The rationale for our approach is prefaced on the assumption that the listed interests are within the LLM's knowledge window.
While this assumption is reasonable for most conventional activities, it is certainly violated for state-of-the-art trends.
However, such a limitation can be easily fixed by providing a description of each activity and computing the embedding on the description, rather than on the name of the activity itself.


This work establishes initial results in utilizing LLMs for multimodal nudging, naturally leading to numerous potential research directions.
For example, while we currently leverage embeddings from various content types in a zero-shot learning context, these embeddings could also serve as input features for additional learning via e.g. a reinforcement learning approach. Furthermore, extending this research to diverse domains like e-commerce, education, and healthcare could uncover fresh avenues for positively shaping user behavior through hyper-personalized multimodal nudging applications.

\bibliographystyle{abbrv}
\bibliography{references}

\begin{thebibliography}{10}

\bibitem{dalecke2020designing}
S.~Dalecke and R.~Karlsen.
\newblock Designing dynamic and personalized nudges.
\newblock In {\em Proceedings of the 10th International Conference on Web
  Intelligence, Mining and Semantics}, pages 139--148, 2020.

\bibitem{davies2012associations}
C.~A. Davies, C.~Vandelanotte, M.~J. Duncan, and J.~G. van Uffelen.
\newblock Associations of physical activity and screen-time on health related
  quality of life in adults.
\newblock {\em Preventive medicine}, 55(1):46--49, 2012.

\bibitem{de2011sedentary}
L.~de~Wit, A.~van Straten, F.~Lamers, P.~Cuijpers, and B.~Penninx.
\newblock Are sedentary television watching and computer use behaviors
  associated with anxiety and depressive disorders?
\newblock {\em Psychiatry research}, 186(2-3):239--243, 2011.

\bibitem{dereventsov2022simulated}
A.~Dereventsov and A.~Bibin.
\newblock Simulated contextual bandits for personalization tasks from
  recommendation datasets.
\newblock In {\em 2022 IEEE International Conference on Data Mining Workshops
  (ICDMW)}, pages 1--6. IEEE, 2022.

\bibitem{dereventsov2021unreasonable}
A.~Dereventsov, R.~Vatsavai, and C.~G. Webster.
\newblock On the unreasonable efficiency of state space clustering in
  personalization tasks.
\newblock In {\em 2021 International Conference on Data Mining Workshops
  (ICDMW)}, pages 742--749. IEEE, 2021.

\bibitem{ding2021zero}
H.~Ding, Y.~Ma, A.~Deoras, Y.~Wang, and H.~Wang.
\newblock Zero-shot recommender systems.
\newblock {\em arXiv preprint arXiv:2105.08318}, 2021.

\bibitem{fan2023recommender}
W.~Fan, Z.~Zhao, J.~Li, Y.~Liu, X.~Mei, Y.~Wang, J.~Tang, and Q.~Li.
\newblock Recommender systems in the era of large language models (llms).
\newblock {\em arXiv preprint arXiv:2307.02046}, 2023.

\bibitem{gomez2022whom}
J.~M. G{\'o}mez~Penedo, B.~Schwartz, J.~Giesemann, J.~A. Rubel, A.-K.
  Deisenhofer, and W.~Lutz.
\newblock For whom should psychotherapy focus on problem coping? a machine
  learning algorithm for treatment personalization.
\newblock {\em Psychotherapy Research}, 32(2):151--164, 2022.

\bibitem{hamer2013television}
M.~Hamer, L.~Poole, and N.~Messerli-B{\"u}rgy.
\newblock Television viewing, c-reactive protein, and depressive symptoms in
  older adults.
\newblock {\em Brain, behavior, and immunity}, 33:29--32, 2013.

\bibitem{hassouni2018personalization}
A.~e. Hassouni, M.~Hoogendoorn, M.~v. Otterlo, and E.~Barbaro.
\newblock Personalization of health interventions using cluster-based
  reinforcement learning.
\newblock In {\em International Conference on Principles and Practice of
  Multi-Agent Systems}, pages 467--475. Springer, 2018.

\bibitem{hou2023large}
Y.~Hou, J.~Zhang, Z.~Lin, H.~Lu, R.~Xie, J.~McAuley, and W.~X. Zhao.
\newblock Large language models are zero-shot rankers for recommender systems.
\newblock {\em arXiv preprint arXiv:2305.08845}, 2023.

\bibitem{hu2003television}
F.~B. Hu, T.~Y. Li, G.~A. Colditz, W.~C. Willett, and J.~E. Manson.
\newblock Television watching and other sedentary behaviors in relation to risk
  of obesity and type 2 diabetes mellitus in women.
\newblock {\em Jama}, 289(14):1785--1791, 2003.

\bibitem{ie2019recsim}
E.~Ie, C.-w. Hsu, M.~Mladenov, V.~Jain, S.~Narvekar, J.~Wang, R.~Wu, and
  C.~Boutilier.
\newblock Recsim: A configurable simulation platform for recommender systems.
\newblock {\em arXiv preprint arXiv:1909.04847}, 2019.

\bibitem{jesse2021digital}
M.~Jesse and D.~Jannach.
\newblock Digital nudging with recommender systems: Survey and future
  directions.
\newblock {\em Computers in Human Behavior Reports}, 3:100052, 2021.

\bibitem{karlsen2019recommendations}
R.~Karlsen and A.~Andersen.
\newblock Recommendations with a nudge.
\newblock {\em Technologies}, 7(2):45, 2019.

\bibitem{datareportal2022}
{Kemp}.
\newblock Digital 2022: Global overview report.
\newblock
  \url{https://datareportal.com/reports/digital-2022-global-overview-report},
  2022.
\newblock Accessed: 2023-08-18.

\bibitem{lex2021psychology}
E.~Lex, D.~Kowald, P.~Seitlinger, T.~N.~T. Tran, A.~Felfernig, M.~Schedl,
  et~al.
\newblock Psychology-informed recommender systems.
\newblock {\em Foundations and Trends{\textregistered} in Information
  Retrieval}, 15(2):134--242, 2021.

\bibitem{li2019zero}
J.~Li, M.~Jing, K.~Lu, L.~Zhu, Y.~Yang, and Z.~Huang.
\newblock From zero-shot learning to cold-start recommendation.
\newblock In {\em Proceedings of the AAAI conference on artificial
  intelligence}, volume~33, pages 4189--4196, 2019.

\bibitem{li2023blip}
J.~Li, D.~Li, S.~Savarese, and S.~Hoi.
\newblock Blip-2: Bootstrapping language-image pre-training with frozen image
  encoders and large language models.
\newblock {\em arXiv preprint arXiv:2301.12597}, 2023.

\bibitem{lin2023can}
J.~Lin, X.~Dai, Y.~Xi, W.~Liu, B.~Chen, X.~Li, C.~Zhu, H.~Guo, Y.~Yu, R.~Tang,
  et~al.
\newblock How can recommender systems benefit from large language models: A
  survey.
\newblock {\em arXiv preprint arXiv:2306.05817}, 2023.

\bibitem{liu2023multimodal}
Q.~Liu, J.~Hu, Y.~Xiao, J.~Gao, and X.~Zhao.
\newblock Multimodal recommender systems: A survey.
\newblock {\em arXiv preprint arXiv:2302.03883}, 2023.

\bibitem{madhav2017association}
K.~Madhav, S.~P. Sherchand, and S.~Sherchan.
\newblock Association between screen time and depression among us adults.
\newblock {\em Preventive medicine reports}, 8:67--71, 2017.

\bibitem{middleton2013long}
K.~R. Middleton, S.~D. Anton, and M.~G. Perri.
\newblock Long-term adherence to health behavior change.
\newblock {\em American journal of lifestyle medicine}, 7(6):395--404, 2013.

\bibitem{nabizadeh2020learning}
A.~H. Nabizadeh, J.~P. Leal, H.~N. Rafsanjani, and R.~R. Shah.
\newblock Learning path personalization and recommendation methods: A survey of
  the state-of-the-art.
\newblock {\em Expert Systems with Applications}, 159:113596, 2020.

\bibitem{portugal2018use}
I.~Portugal, P.~Alencar, and D.~Cowan.
\newblock The use of machine learning algorithms in recommender systems: A
  systematic review.
\newblock {\em Expert Systems with Applications}, 97:205--227, 2018.

\bibitem{pourpanah2022review}
F.~Pourpanah, M.~Abdar, Y.~Luo, X.~Zhou, R.~Wang, C.~P. Lim, X.-Z. Wang, and
  Q.~J. Wu.
\newblock A review of generalized zero-shot learning methods.
\newblock {\em IEEE transactions on pattern analysis and machine intelligence},
  2022.

\bibitem{rohde2018recogym}
D.~Rohde, S.~Bonner, T.~Dunlop, F.~Vasile, and A.~Karatzoglou.
\newblock Recogym: A reinforcement learning environment for the problem of
  product recommendation in online advertising.
\newblock {\em arXiv preprint arXiv:1808.00720}, 2018.

\bibitem{salah2020cornac}
A.~Salah, Q.-T. Truong, and H.~W. Lauw.
\newblock Cornac: A comparative framework for multimodal recommender systems.
\newblock {\em The Journal of Machine Learning Research}, 21(1):3803--3807,
  2020.

\bibitem{shmakov2021nudge}
A.~Shmakov.
\newblock Nudge in the conditions of digital transformation: Behavioral basis.
\newblock {\em Journal of Institutional Studies}, 13(3):102--116, 2021.

\bibitem{su2009survey}
X.~Su and T.~M. Khoshgoftaar.
\newblock A survey of collaborative filtering techniques.
\newblock {\em Advances in artificial intelligence}, 2009, 2009.

\bibitem{sutton2018reinforcement}
R.~S. Sutton and A.~G. Barto.
\newblock {\em Reinforcement learning: An introduction}.
\newblock MIT press, 2018.

\bibitem{thaler2009nudge}
R.~H. Thaler and C.~R. Sunstein.
\newblock {\em Nudge: Improving decisions about health, wealth, and happiness}.
\newblock Penguin, 2009.

\bibitem{truong2019multimodal}
Q.-T. Truong and H.~Lauw.
\newblock Multimodal review generation for recommender systems.
\newblock In {\em The World Wide Web Conference}, pages 1864--1874, 2019.

\bibitem{twenge2019more}
J.~M. Twenge.
\newblock More time on technology, less happiness? associations between
  digital-media use and psychological well-being.
\newblock {\em Current Directions in Psychological Science}, 28(4):372--379,
  2019.

\bibitem{vlaev2016theory}
I.~Vlaev, D.~King, P.~Dolan, and A.~Darzi.
\newblock The theory and practice of “nudging”: changing health behaviors.
\newblock {\em Public Administration Review}, 76(4):550--561, 2016.

\bibitem{wang2023zero}
L.~Wang and E.-P. Lim.
\newblock Zero-shot next-item recommendation using large pretrained language
  models.
\newblock {\em arXiv preprint arXiv:2304.03153}, 2023.

\bibitem{wang2019survey}
W.~Wang, V.~W. Zheng, H.~Yu, and C.~Miao.
\newblock A survey of zero-shot learning: Settings, methods, and applications.
\newblock {\em ACM Transactions on Intelligent Systems and Technology (TIST)},
  10(2):1--37, 2019.

\bibitem{wu2021mm}
C.~Wu, F.~Wu, T.~Qi, and Y.~Huang.
\newblock Mm-rec: multimodal news recommendation.
\newblock {\em arXiv preprint arXiv:2104.07407}, 2021.

\bibitem{zhou2023comprehensive}
H.~Zhou, X.~Zhou, Z.~Zeng, L.~Zhang, and Z.~Shen.
\newblock A comprehensive survey on multimodal recommender systems: Taxonomy,
  evaluation, and future directions.
\newblock {\em arXiv preprint arXiv:2302.04473}, 2023.

\bibitem{zhu2018robust}
F.~Zhu, J.~Guo, R.~Li, and J.~Huang.
\newblock Robust actor-critic contextual bandit for mobile health (mhealth)
  interventions.
\newblock In {\em Proceedings of the 2018 acm international conference on
  bioinformatics, computational biology, and health informatics}, pages
  492--501, 2018.

\end{thebibliography}

\clearpage
\newgeometry{top=.4in, bottom=0in, left=.65in, right=.65in}
\pagestyle{empty}
\onecolumn
\appendix

\begin{figure}[t]
    \centering
    \vspace*{-.5in}
    \includegraphics[width=.9\linewidth]{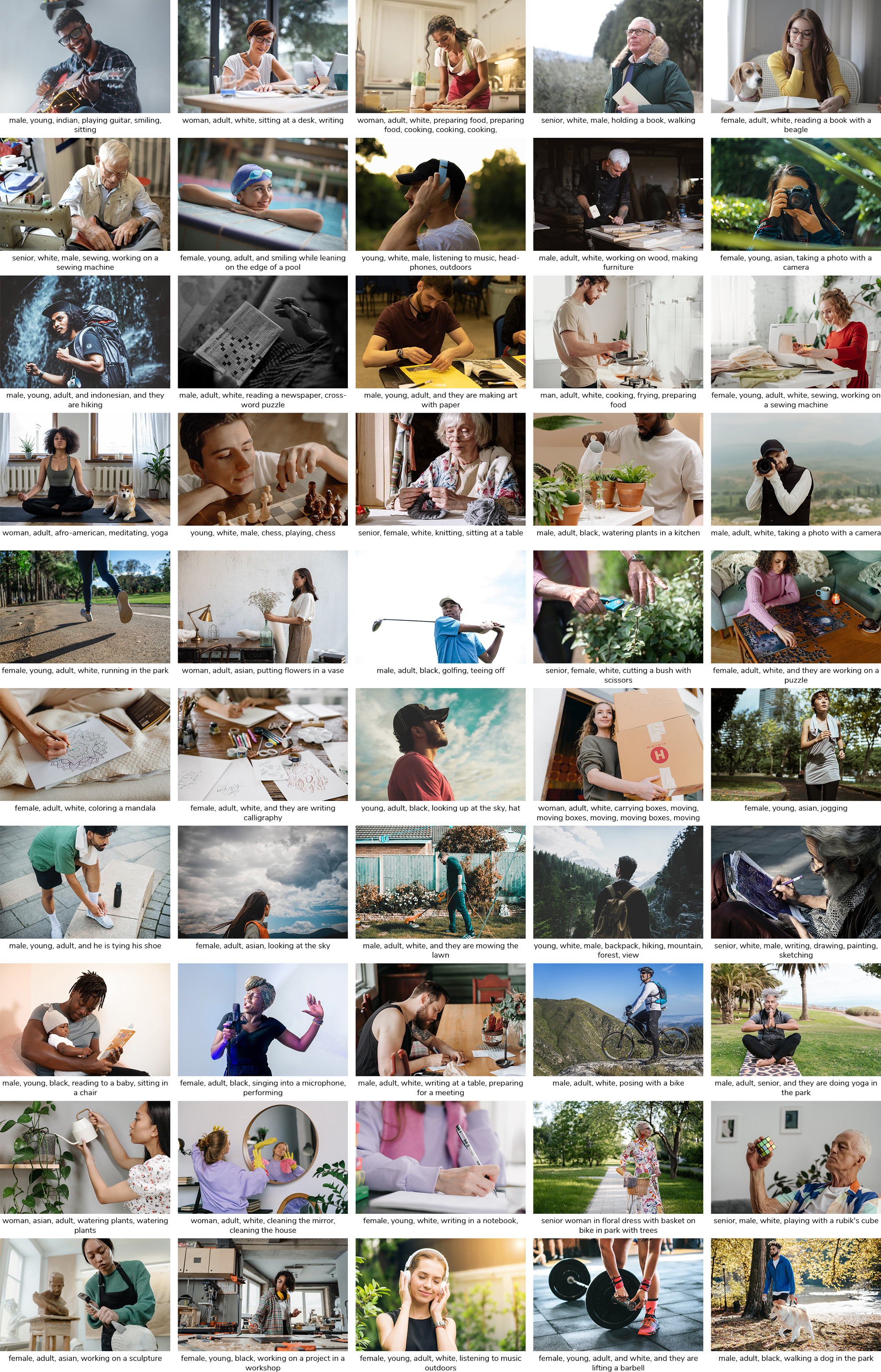}
    \caption{Images selected from Pexels and the corresponding captions generated via the BLIP-2 framework.}
    \label{fig:generated_images_full}
\end{figure}

This appendix contains the entire dataset generated for the synthetic screen time monitoring application, as explained in Section~\ref{sec:problem_setting}.
Specifically, Table~\ref{tab:generated_users_full} contains all 20 users, Table~\ref{tab:generated_messages_full} contains all 40 messages, and Figure~\ref{fig:generated_images_full} contains the 50 selected images and their corresponding captions (see also Table~\ref{tab:generated_captions} for captions alone).

The full dataset and the source code can be found at~\url{https://github.com/paxnea/LLM-multimodal-nudging}.

\begin{table}[ht]
    \centering\fontsize{7.7}{7.7}\selectfont
    \caption{Users generated via GPT-4.}
    \label{tab:generated_users_full}
    \begin{filecontents*}{./data/users.csv}
,Gender,Age,Race,Likes,Dislikes
0,Female,23,White,{active, outdoors, learning},{mental, crafts, homemaking}
1,Male,42,Black,{outdoors, mental, active},{learning, exploration, arts}
2,Female,25,Asian,{crafts, indoors, arts},{passive, mental, outdoors}
3,Male,68,Hispanic,{passive, indoors, homemaking},{relaxation, learning, physical}
4,Female,39,White,{mental, homemaking, crafts},{outdoors, active, relaxation}
5,Male,43,Black,{outdoors, exploration, relaxation},{indoors, passive, homemaking}
6,Female,22,Hispanic,{active, mental, homenaking},{learning, exploration, relaxation}
7,Male,45,Asian,{indoors, passive, exploration},{outdoors, active, mental}
8,Female,35,White,{active, outdoors, exploration},{passive, indoors, physical}
9,Male,31,Hispanic,{mental, learning, relaxation},{outdoors, active, arts}
10,Female,26,Black,{indoors, passive, crafts},{outdoors, active, mental}
11,Male,39,Asian,{active, mental, homenaking},{learning, exploration, relaxation}
12,Female,40,White,{indoors, passive, exploration},{outdoors, active, mental}
13,Male,49,Hispanic,{outdoors, active, arts},{passive, indoors, physical}
14,Female,37,Black,{mental, learning, crafts},{outdoors, active, relaxation}
15,Male,41,White,{outdoors, active, exploration},{passive, indoors, physical}
16,Female,48,Asian,{active, mental, homenaking},{learning, exploration, relaxation}
17,Male,56,Hispanic,{indoors, passive, exploration},{outdoors, active, mental}
18,Female,53,White,{outdoors, active, arts},{passive, indoors, physical}
19,Male,60,Black,{mental, learning, crafts},{outdoors, active, relaxation}
\end{filecontents*}
\csvautobooktabular{./data/users.csv}
    \vspace*{5ex}
    \centering\fontsize{7.7}{7.7}\selectfont
    \caption{Captions generated via BLIP-2.}
    \label{tab:generated_captions}
    \begin{filecontents*}{./data/images.csv}
,Caption
0,{male, young, indian, playing guitar, smiling, sitting}
1,{woman, adult, white, sitting at a desk, writing}
2,{woman, adult, white, preparing food, preparing food, cooking, cooking, cooking,}
3,{senior, white, male, holding a book, walking}
4,{female, adult, white, reading a book with a beagle}
5,{senior, white, male, sewing, working on a sewing machine}
6,{female, young, adult, and smiling while leaning on the edge of a pool}
7,{young, white, male, listening to music, headphones, outdoors}
8,{male, adult, white, working on wood, making furniture}
9,{female, young, asian, taking a photo with a camera}
10,{male, young, adult, and indonesian, and they are hiking}
11,{male, adult, white, reading a newspaper, crossword puzzle}
12,{male, young, adult, and they are making art with paper}
13,{man, adult, white, cooking, frying, preparing food}
14,{female, young, adult, white, sewing, working on a sewing machine}
15,{woman, adult, afro-american, meditating, yoga}
16,{young, white, male, chess, playing, chess}
17,{senior, female, white, knitting, sitting at a table}
18,{male, adult, black, watering plants in a kitchen}
19,{male, adult, white, taking a photo with a camera}
20,{female, young, adult, white, running in the park}
21,{woman, adult, asian, putting flowers in a vase}
22,{male, adult, black, golfing, teeing off}
23,{senior, female, white, cutting a bush with scissors}
24,{female, adult, white, and they are working on a puzzle}
25,{female, adult, white, coloring a mandala}
26,{female, adult, white, and they are writing calligraphy}
27,{young, adult, black, looking up at the sky, hat}
28,{woman, adult, white, carrying boxes, moving, moving boxes, moving, moving boxes, moving}
29,{female, young, asian, jogging}
30,{male, young, adult, and he is tying his shoe}
31,{female, adult, asian, looking at the sky}
32,{male, adult, white, and they are mowing the lawn}
33,{young, white, male, backpack, hiking, mountain, forest, view}
34,{senior, white, male, writing, drawing, painting, sketching}
35,{male, young, black, reading to a baby, sitting in a chair}
36,{female, adult, black, singing into a microphone, performing}
37,{male, adult, white, writing at a table, preparing for a meeting}
38,{male, adult, white, posing with a bike}
39,{male, adult, senior, and they are doing yoga in the park}
40,{woman, asian, adult, watering plants, watering plants}
41,{woman, adult, white, cleaning the mirror, cleaning the house}
42,{female, young, white, writing in a notebook,}
43,{senior woman in floral dress with basket on bike in park with trees}
44,{senior, male, white, playing with a rubik's cube}
45,{female, adult, asian, working on a sculpture}
46,{female, young, black, working on a project in a workshop}
47,{female, young, adult, white, listening to music outdoors}
48,{female, young, adult, and white, and they are lifting a barbell}
49,{male, adult, black, walking a dog in the park}
\end{filecontents*}
\csvautobooktabular{./data/images.csv}
\end{table}

\begin{table}[t]
    \vspace*{-1in}
    \centering\fontsize{8}{8}\selectfont
    \caption{Messages generated via GPT-3.5.}
    \label{tab:generated_messages_full}
    \begin{filecontents*}{./data/messages.csv}
,Message
0,{Get up and groove! Dancing is your ticket to fun, fitness, and a break from the screen. Let the music move you!}
1,{Ditch the screen, hit the trail! Discover nature's wonders, boost your mood, and energize your body with an invigorating hike.}
2,{Give your eyes a break and let your body flow with the ancient practice of yoga. Feel the tension melt away and experience the incredible benefits of increased flexibility, strength, and inner peace.}
3,{Hey! Why not leave the screen behind and hop on your bike? Feel the wind in your hair, soak up the sunshine, and boost your mood while getting some exercise. Let's go biking and enjoy the great outdoors!}
4,{Step away from the screen and let's stroll! Walking boosts energy, clears the mind, and fills your day with fresh inspiration. Let's get moving!}
5,{Dive into the pool and feel the freedom! Swap screens for a refreshing swim, unleash your energy, and let the water ignite your senses.}
6,{Why not give your thumbs a break and start pumping some iron? Weightlifting will boost your strength, tone your muscles, and leave you feeling unstoppable!}
7,{Hey! Ready to get moving? Put down the screen and let's hit the pavement - jogging boosts energy, reduces stress, and gives you a natural high! Let's go!}
8,{Step away from the screen and let's roll! Experience the thrill of roller skating - it's exhilarating, boosts your mood, and keeps you active. Lace up and let the fun begin!}
9,{Step away from the screen and tee off into a world of fun and relaxation! Golfing boosts your mood, improves focus, and gets you outdoors enjoying nature. Let's swing into action!}
10,{Why not give your brain a boost and challenge yourself with Sudoku? It's a fun way to sharpen your mind and improve your problem-solving skills!}
11,{Give your eyes a break and let your imagination soar with the magic of reading. Rediscover the joy of stories, expand your knowledge, and unlock endless possibilities through the power of books.}
12,{Why not give your brain a workout and have some fun at the same time? Grab a jigsaw puzzle and let your problem-solving skills shine while enjoying the benefits of improved focus and relaxation!}
13,{Why not challenge your brain and have some fun with crosswords? They boost your mental agility and memory while giving you a break from screens!}
14,{Why not discover new horizons and boost your brainpower by immersing yourself in the captivating world of language learning? Open the door to endless possibilities and embrace the incredible benefits of expanding your linguistic skills!}
15,{Give your mind a well-deserved vacation and unlock a world of calm and clarity through the power of meditation. Experience the transformative benefits that await you!}
16,{Step away from the screen and dive into the captivating world of chess - sharpen your focus, boost problem-solving skills, and have a blast strategizing your way to victory!}
17,{Feeling screen overload? Time to conquer the colorful challenge of a Rubik's cube! Unleash your problem-solving skills, improve focus, and enjoy a satisfying victory! Let's cube it!}
18,{Why stare at screens when you can immerse yourself in the incredible world of music? Let the captivating melodies transport you, boost your mood, and ignite your imagination!}
19,{Why stare at screens when you can conquer challenging card layouts, boost your focus, and relax with a satisfying game of solitaire? Give it a try, you'll be amazed!}
20,{Why not add some color to your day? Put down the screen and let your imagination soar as you relax and find inner calm through the therapeutic art of coloring.}
21,{Give your eyes a break and let your imagination soar! Grab a pen and paper, and discover the joy of drawing while boosting your creativity and reducing screen time.}
22,{Why not turn off the screen and embark on a delightful journey of preserving memories with scrapbooking - spark joy, unleash your creativity, and create something beautiful to cherish forever!}
23,{Why not let your imagination soar and bring your stories to life through the power of creative writing? It's a perfect way to unwind, express yourself, and unlock a world of endless possibilities!}
24,{Give your eyes a break and let your voice shine! Singing boosts mood, relieves stress, and brings joy to your day. So put down the screen and start serenading!}
25,{Why stare at a screen when you can create stunning art with a pen? Embrace the beauty and mindfulness of calligraphy, and unlock your artistic potential!}
26,{Why not switch off your screen and dive into the world of embroidery? It's a fantastic way to relax, unleash your creativity, and create something beautiful!}
27,{Step away from your screen and let your hands mold something extraordinary. Sculpting is a creative adventure that unleashes your imagination and brings joy to your soul. Start sculpting now and experience the incredible benefits of self-expression and relaxation.}
28,{Capture the world through your lens and unlock a world of creativity, beauty, and self-expression. Embrace the magic of photography and watch your screen time fade away!}
29,{Why not turn off the screen and let your creativity flow through the magic of songwriting? Unleash your inner artist and experience the joy of self-expression while reaping the benefits of relaxation and inspiration.}
30,{Give your screen a break and let's clean instead! Cleaning not only tidies up your space, but also clears your mind and brings a fresh start to your day. Let's get cleaning and enjoy the benefits of a clean and organized space!}
31,{Why not whip up a delicious meal and discover the joy of creating something amazing in the kitchen? Cooking can be a tasty way to unwind, explore new flavors, and nourish your body and soul.}
32,{Why not give your eyes a break and pick up some knitting needles instead? Knitting is a relaxing and rewarding hobby that can help reduce screen time and boost your creativity!}
33,{Put down the screen and pick up a needle - sewing is waiting to unleash your creativity and bring a sense of fulfillment and joy to your day!}
34,{Take a break from screens and discover the joy of gardening - it's like a breath of fresh air, boosts your mood, and lets your inner green thumb bloom!}
35,{Get off your screen and embrace the joy of woodworking! Discover the satisfaction of creating something with your own hands while unleashing your imagination and reducing stress. Start your woodworking adventure today!}
36,{Why not give your eyes a break and whip up something delicious in the kitchen? Baking is a tasty way to relax, get creative, and treat yourself!}
37,{Why not give your screen a rest and transform your space into a cozy oasis? Discover the joy of home decorating and create a sanctuary that brings you calm and happiness.}
38,{Why not trade your screen for a scenic green? Get outside, breathe in fresh air, and transform your surroundings with the beauty and tranquility of landscaping.}
39,{Why not give your eyes a break and dive into the satisfying world of decluttering? Clearing your space not only brings a sense of calm, but also boosts productivity and sparks creativity!}
\end{filecontents*}
\csvreader[tabular=lp{.9\linewidth},
               table head=\toprule&Message\\\midrule,
               late after last line=\\\bottomrule]
        {./data/messages.csv}{}{\csvcoli&\csvcolii}
\end{table}

\end{document}